\documentclass[conference]{IEEEtran}
\IEEEoverridecommandlockouts
\usepackage{graphicx}
\ifCLASSINFOpdf
\else
\fi
\begin{document}
%
\title{Inflection system of a language as a complex network}

\author{\IEEEauthorblockN{Henryk Fuk\'s \thanks{978-1-4244-3878-5/09/\$25.00 \textcopyright2009 IEEE}}
\IEEEauthorblockA{Department of Mathematics\\
Brock University\\
St. Catharines, ON, Canada\\
Email: hfuks@brocku.ca}
}
\maketitle

\begin{abstract}
We investigate inflection structure of a synthetic
language using  Latin as an example. We construct a bipartite graph 
in which one group of vertices correspond to dictionary headwords
and the other group to inflected forms encountered in a given text.
Each inflected form is connected to its corresponding headword,
which in some cases in non-unique. The resulting sparse graph
decomposes into a large number of connected components, to
be called word groups. We then show how the concept of the word
group can be used to construct coverage curves of selected Latin
texts. We also investigate a version of the inflection graph in
which all theoretically possible inflected forms are included. Distribution
of sizes of connected components of this graphs resembles 
cluster distribution in a lattice percolation near the critical point.

\end{abstract}

\section{Introduction}
Vocabulary of human languages can be viewed as a large and complex network or graph,
in which individual vertices represent words or families of words, and edges represent
relationships between words. Many such models have been studied in recent years,
including  networks of co-occurrences of words in sentences \cite{Ferrer2001},
thesaurus graphs \cite{MotterdLD02,KinouchiMLLR02,HolandaPKMR04}, WordNet database graphs \cite{citeulike:1179006},
and many others   \cite{KeY08,CaldeiraLANM06,PomiM04,CanchoSK04,AntiqueiraNOC07}.

Much of the aforementioned work has been done in the context of the English language, which,
among other characteristic properies, exhibits  only a minimal inflection, especially if compared to
other Indo-European languages. In analytic languages like English, grammatical categories
and relations are handled mostly by the word order, and not by the inflection. 

In contrast to this, synthetic languages such as Latin, Greek, Polish, or Russian make an extensive use 
of inflection, and one word in these languages can appear in great many forms. 
reflecting grammatical categories such as tense, mood, person, number, gender, case, etc.
In the past, there was relatively little work done on modelling of inflected languages using the
paradigm of complex networks, and the goal of this  paper is to present some
initial findings of the author in this area.

Given the abundance of synthetic languages, one faces the issue of selecting
one of them for detailed analysis. The language which has been most heavily studied
and for which the largest body of literature exists is a natural choice  -- and there is no
doubt  that Latin must be chosen given these criteria. Latin literature stretches for
over 20 centuries ranging from the literature of ancient Rome right up to the
21st century. Latin served as  \textit{lingua franca} for Western civilization for many
centuries, so there is no shortage of Latin texts, and a vast number of them is available in
electronic form.

\begin{figure}
\centering
\includegraphics[width=3.0in]{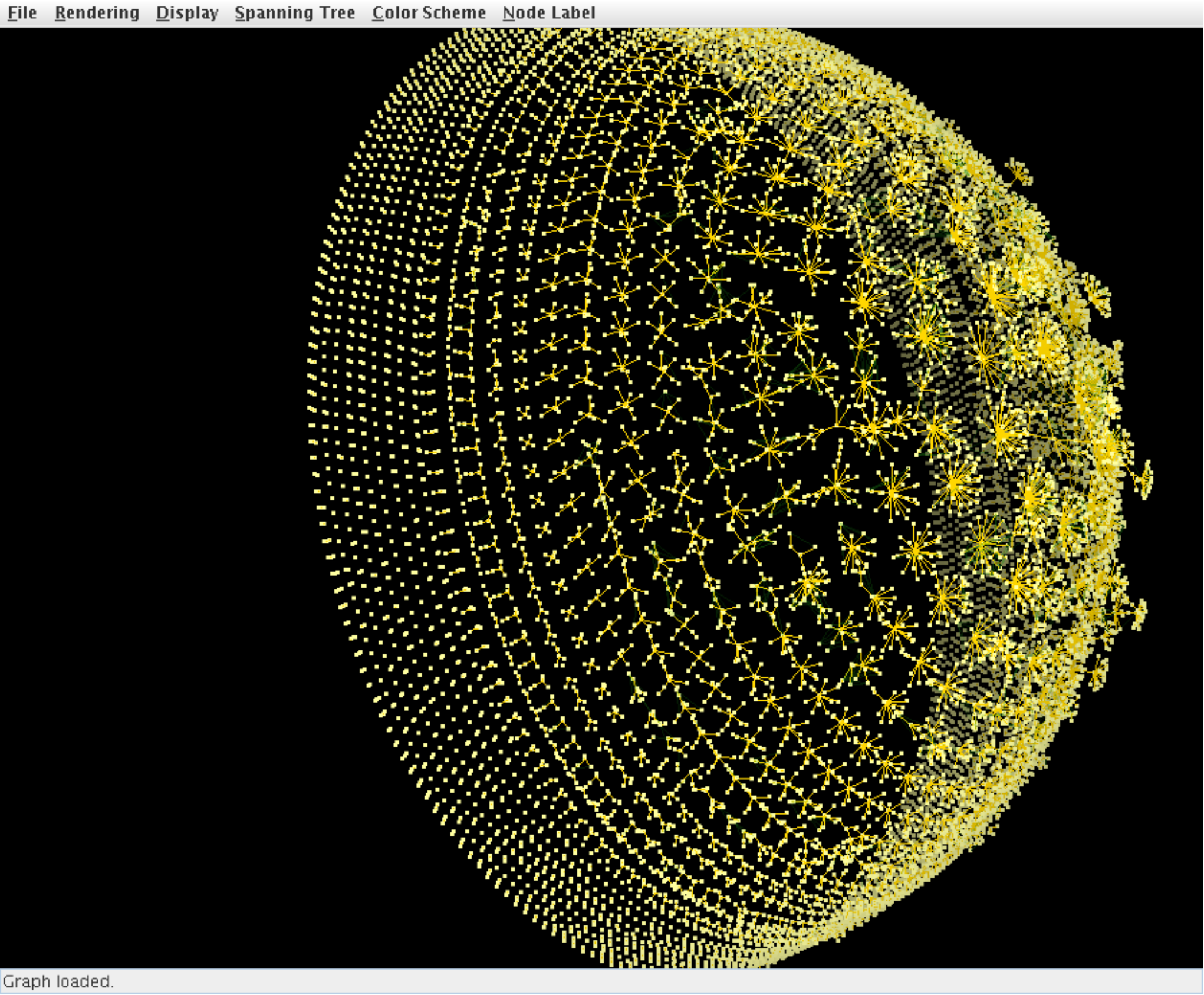}
\caption{Visualization of the inflection graph 
of \textit{De bello Gallico}.}
\label{gallvis}
\end{figure}

In addition to this, Latin inflectional system has been studied so extensively that
literally every single aspect of this system is exceptionally well documented.
Software tools which ``understand''  Latin inflection system are also readily available,
including an excellent open-source WORDS program written by W. Whitaker
\cite{words}.

\section{Motivation}
One of the motivation of this work was the problem of vocabulary size. Let us suppose
that we want to count how many distinct words a given work contains -- for example, for the
purpose of comparing two works and deciding which one is more ``difficult'' as long
as vocabulary is concerned. How do we do this in a language like Latin, where 
one dictionary headword can have as many as hundreds of different forms? In addition
to this, in some cases, one inflectional form can correspond to more than one dictionary headword,
and one must deduce from the contexts which one to choose.

The earliest qualitative approach to Latin vocabulary can be found in the Ph.D. thesis
of Paul Bernard Diederich \cite{Diederich39} from 1939, who performed a count of Latin
headwords occurring in a selection of texts from 200 Latin authors totaling over
200,000 words. He did this entirely by hand -- a hardly attractive proposition 
in todays computerized world. 

If one wants to perform computerized count of words, and wants to count various inflectional forms
of the same headword as one entry, one has to understand the relationship between inflected 
forms and headwords. This can be done by introducting the concept of an \textit{inflection graph}.

\section{Inflection graph}
The inflection graph for a given text is a bipartite graph which is constructed as follows.
First, we create a set of vertices, to be denoted by $B$,  corresponding to all distinct
words of the text. I a word occurs in the text more than once, it is
represented by one vertex nevertheless. 
We then go over all vertices in $B$, and check which headwords can possibly
correspond to each of words in $B$. These headwords form another set of vertices,
to be called $A$. In practice, headwords may be obtained by using W. Whitaker's WORDS program. 
As in the case of $B$, elements of $A$ are unique, so that each headword appears in $A$ only
once. 

\begin{figure}
\centering
\includegraphics[width=3.0in]{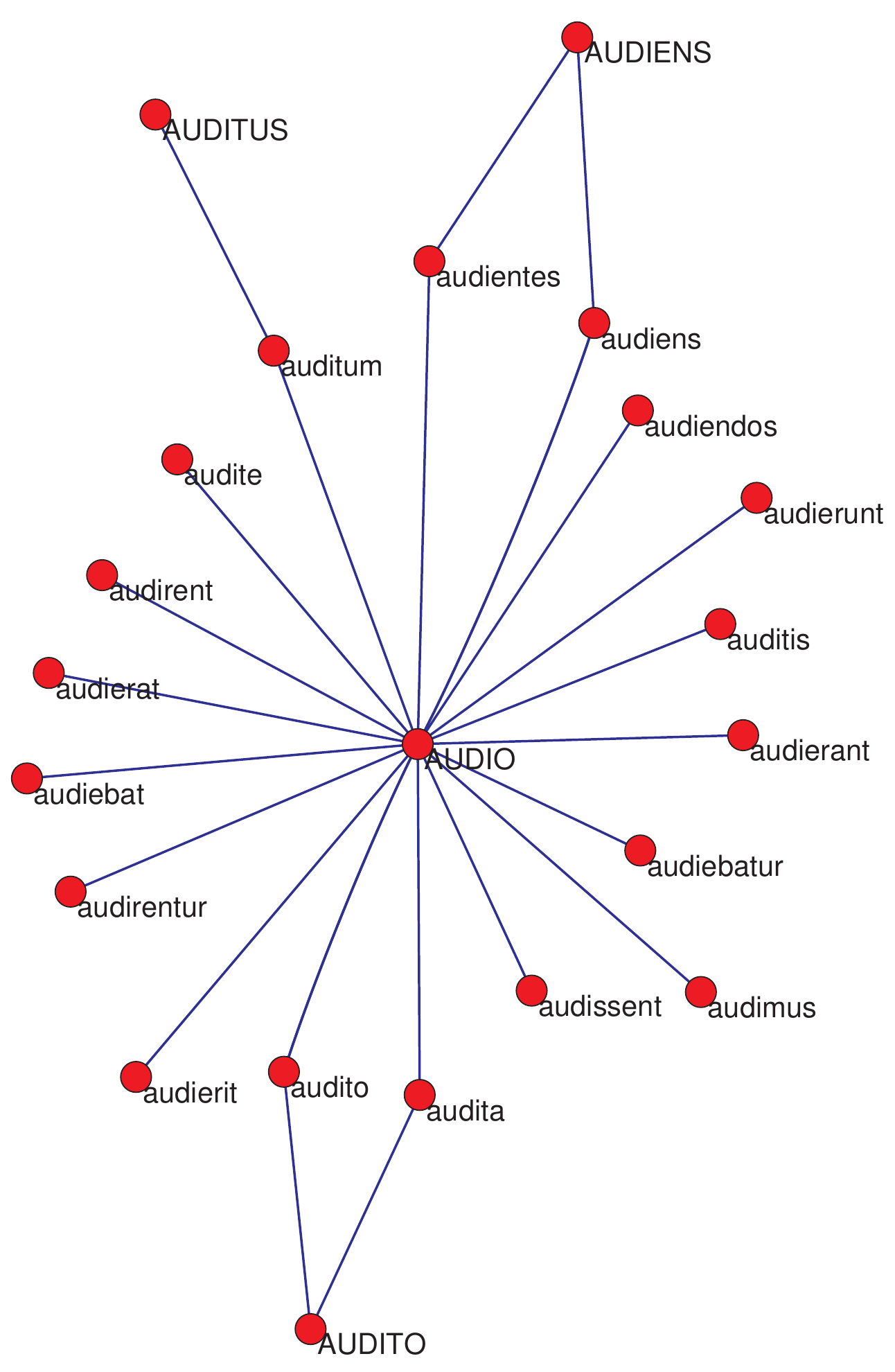}
\caption{Example of a connected component of the inflection graph 
of \textit{De bello Gallico}. Vertices with capitalized labels correspond
to dictionary headwords, and those in lowercase to inflected forms.}
\label{clusterex}
\end{figure}
If an element of $A$ is a headword corresponding to some element of $B$, then these
two are connected. Obviously, for most headwords in $A$, there are many corresponding
inflected forms in $B$, so an element of $A$ is typically connected to many
elements of $B$. For example, \textit{dicunt} (they say) and \textit{dixit} (he said) are both inflected forms
of the verb \textit{dico}, thus we will have a vertex in  $A$ corresponding to \textit{dico}
connected to vertices in $B$ corresponding to \textit{dicunt} (they say) and \textit{dixit}.

 However, the opposite can also be true: in some instances, a word
can be an inflected form of more than one headword, so that elements of $B$ are someties
connected to more than one element of $A$. As an example, consider 
the word \textit{sublatus}, which could be a form of \textit{tollo} (lift, raise) or \textit{suffero} (bear, endure),
thus a vertex of $B$ corresponding to \textit{sublatus} will be connected to vertices
of $A$ corresponding to \textit{tollo}  and \textit{suffero}.

\begin{figure}
\centering
\includegraphics[width=3.5in]{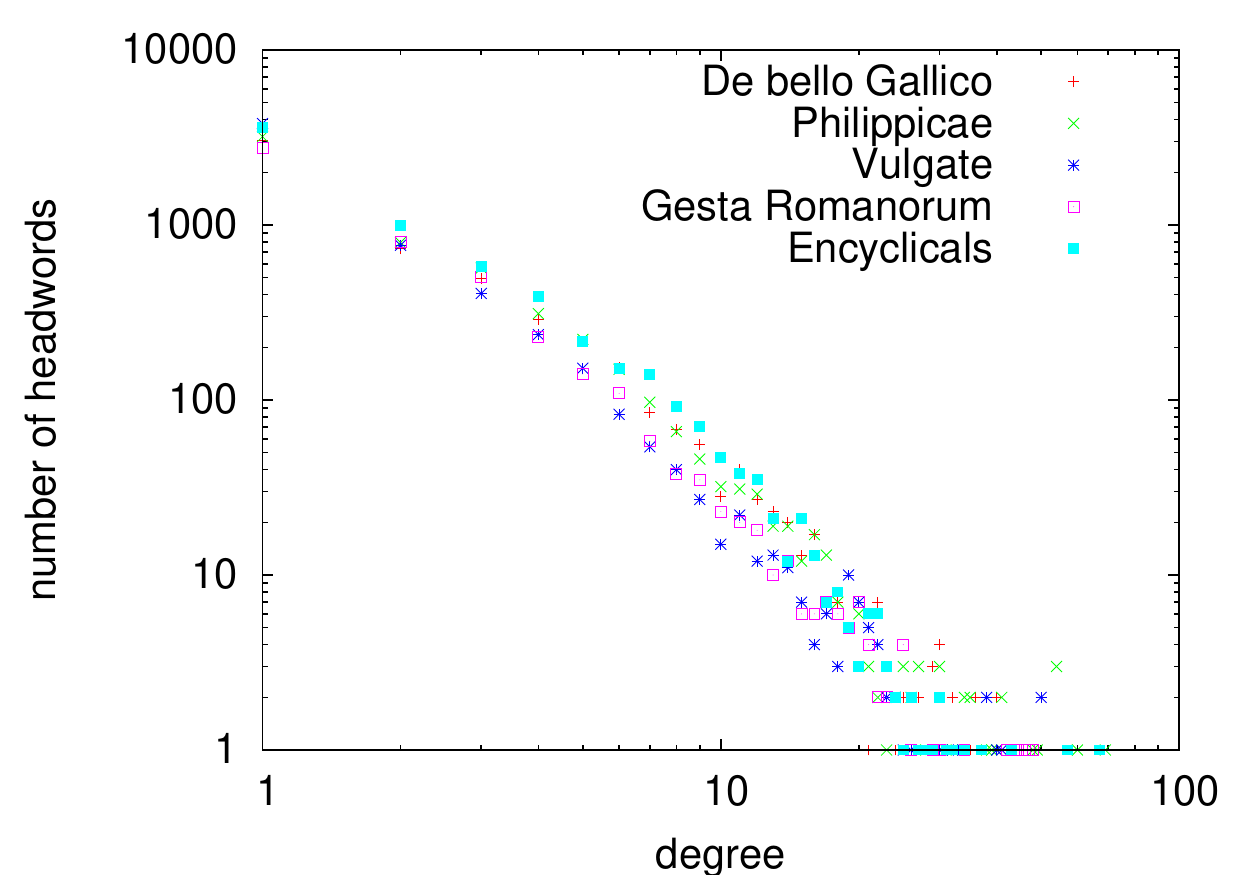}
\caption{Degree distribution of headword vertices.}
\label{head-degdist}
\end{figure}

The bigraph obtained using the aforementioned procedure is typically quite large but not very dense.
For example, for the classic work of Julius Caesar \textit{De bello Gallico} (published in 50s or 40s BC), consisting of
51,300 words,  this bigraph has  5,377 of vertices in $A$, 10,977 vertices in $B$, and 15,349 edges.
Figure~\ref{gallvis} shows a visualization of this graph done by Walrus \cite{walrus}, a software tool for visualizing large  graphs
using 3D hyperbolic geometry and a fisheye-like distortion.
Degree distribution for vertices of type A (headwords) for \textit{De bello Gallico}, as well as some other works
(to be discussed later) is shown in Figure~\ref{head-degdist}.
The distribution seems to have features of a power law. We also observe that the degree of a headword
represents the number of inflected forms of that headword appearing in the text  -- and, as one can see from the
figure, this number can easily approach 100.

\subsection{Connected components}
An important feature of the inflection graph is that it is, obviously, not connected, and that it has a large number
of disjoint components -- in the case of \textit{De bello Gallico}, 3,740 components. Each of these components 
will be called a \textit{word group}. Example of such a component is shown in
Figure~\ref{clusterex}. It consists of four headwords (written in uppercase) and 18 inflected
forms (written in lowercase).

In most cases, words within one group are closely related semantically,
 but not always -- occasionally words with quite different meaning may  belong to the same group,
as in the case of  \textit{tollo} and \textit{suffero} mentioned above.This is especially true for the largest group,
as we will see later on.
 Nevertheless, 
word groups usually closely correspond to what linguists call \textit{word families}. The main advantage 
of word groups lies in  the fact that
they can be easily determined using well known algorithms for computing   connected components of a graph.
We used python package  NetworkX \cite{Networkx} for this purpose. 

 \subsection{Rank-frequency distribution for groups}
\begin{figure}
\centering
\includegraphics[width=3.5in]{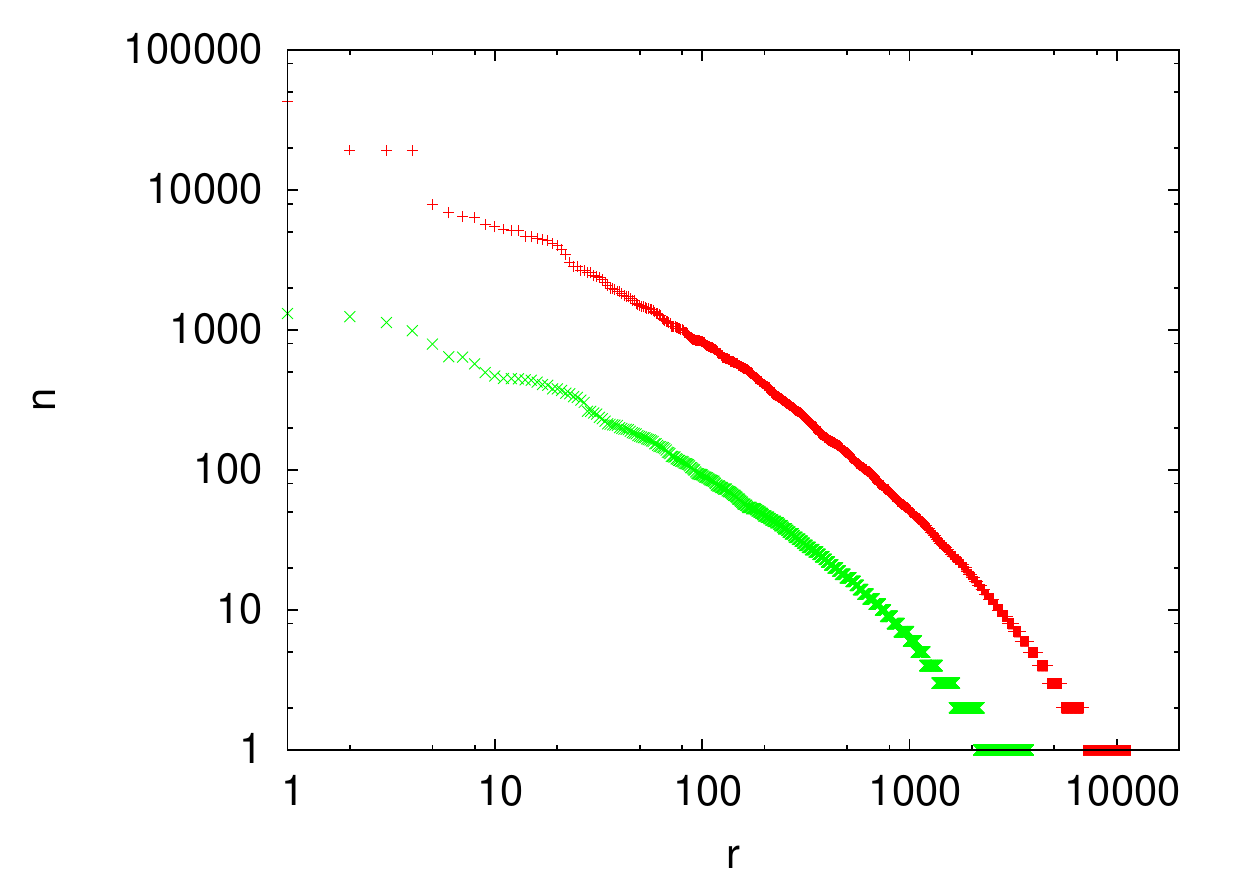}
\caption{Rank-frequency distribution of group occurrences in \textit{Vulgate} ($+$) and \textit{De bello Gallico} ($\times$).}
\label{rankfreqgroups}
\end{figure}

Having the concept of the word group defined, we can now label all groups with
distinct labels, for example, with consecutive integers $i$. If a given word from the text belongs to a group labelled $i$, we will
say that it is an \textit{occurrence of the group} $i$. Obviously, some groups occur more often than others,
so we can sort all groups in decreasing order of occurrences in the text. Position of
a group on this list will be denoted by $r$ (rank), and the number of occurrences of that group
in the text will be denoted by $n_g(r)$. Similar rank-frequency function for individual words
will be denoted by $n_w(r)$.
Figure~\ref{rankfreqgroups} shows log-log plots of $n_g(r)$ versus $r$
for two very different works, namely the aforementioned  \textit{De bello Gallico} and for the Latin Bible translation
of St. Jerome
known as \textit{Vulgate} (AD 390 to 405). In both cases  non-Zipfian behavior is very clear, that is,
the resulting curves a not straight lines.

\subsection{Coverage curves}
A helpful concept used to describe statistical properties of texts is the so called \textit{coverage
curve}. If we assume that the reader of the text knows $k$ top-ranking words, then the text coverage,
or the fraction of known words in the text is defined as
\begin{equation}\label{eqcov}
C_w(k)= \frac{\sum_{r=1}^{k} n_w(r)}{\sum_{r=1}^{N} n_w(r)},
\end{equation}
where $N$ is the total number of distinct words in the text.
The graph of $C_w(k)$ versus $k$ is known as the coverage curve.
Analogously, for groups we can define 
\begin{equation}
C_g(k)= \frac{\sum_{r=1}^{k} n_g(r)}{\sum_{r=1}^{M} n_g(r)},
\end{equation}
where $M$ is the total number of  word groups in the text.

In order to illustrate the difference between  $C_w(k)$ and $C_g(k)$,
we constructed these coverage curves for five different texts.
These texts, in addition to already mentioned \textit{De bello Gallico} 
and \textit{Vulgate}, include Cicero's \textit{Philiphicae} (written 44-43 BC) and
collection of medieval stories \textit{Gesta Romanorum} (13th-14h century).
We also wanted to include some longer contemporary text of considerable
length, which proved to be difficult due to scarcity of such texts.
Finally we somewhat artificially produced a  text by combining two shorter documents.
The resulting file is titled \textit{Encyclicals} and consists of
 two encyclicals of John Paul II,  \textit{Ut Unum Sint}  and
\textit{Evangelium Vitae}. Both of these were issued in the same
year (1995), thus they are sufficiently similar in style to consider  them
as parts of one document. 
Texts of encyclicals were obtained from Vatican repository
\cite{vatican}, and the remaining texts from The Latin Library
\cite{latlib}. In order to make sure that differences in text size
do not interfere with our analysis, all texts have been truncated
so that they have the same length as \textit{De bello Gallico},
that is, 51,300 words.
\begin{figure}
\centering
\includegraphics[width=3.5in]{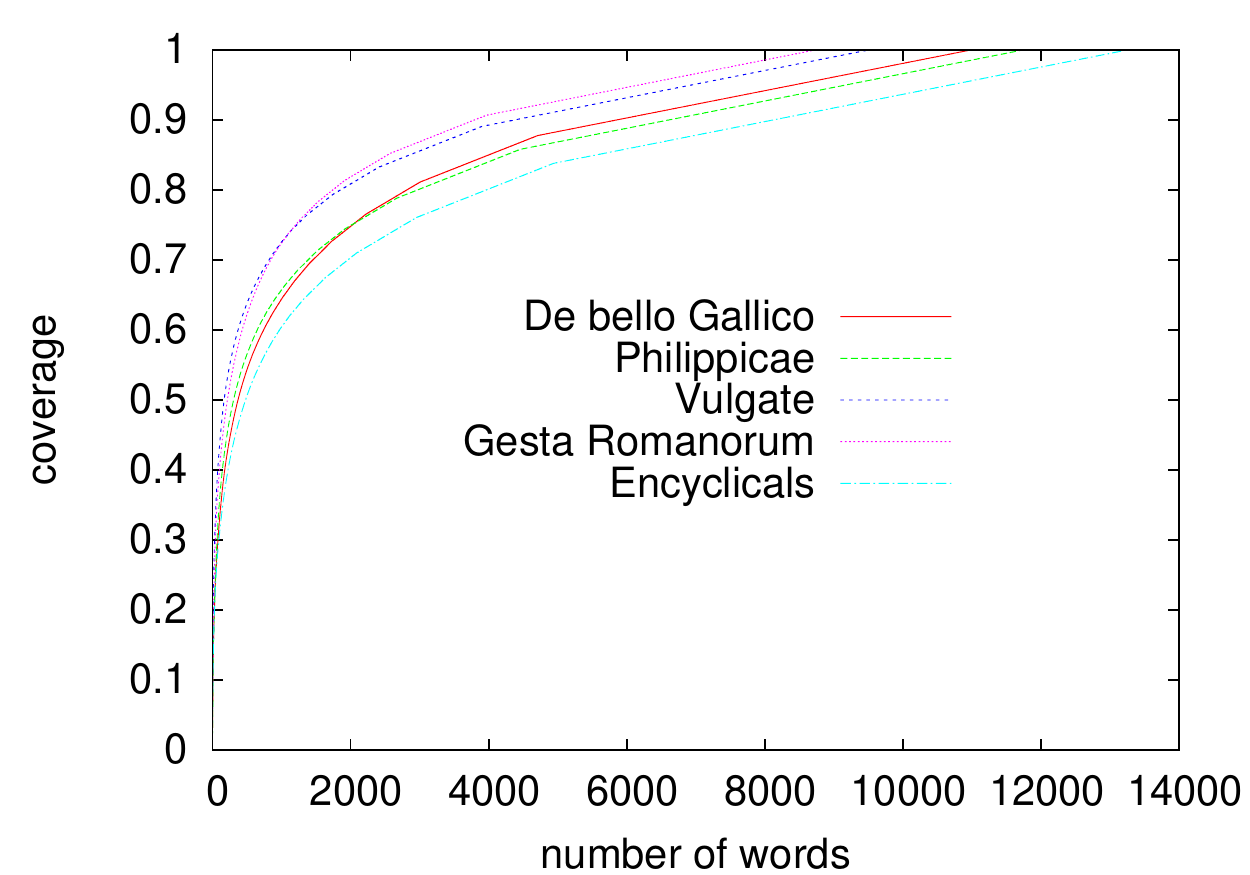}
\caption{Coverage curves for words.}
\label{covwords}
\end{figure}
\begin{figure}
\centering
\includegraphics[width=3.5in]{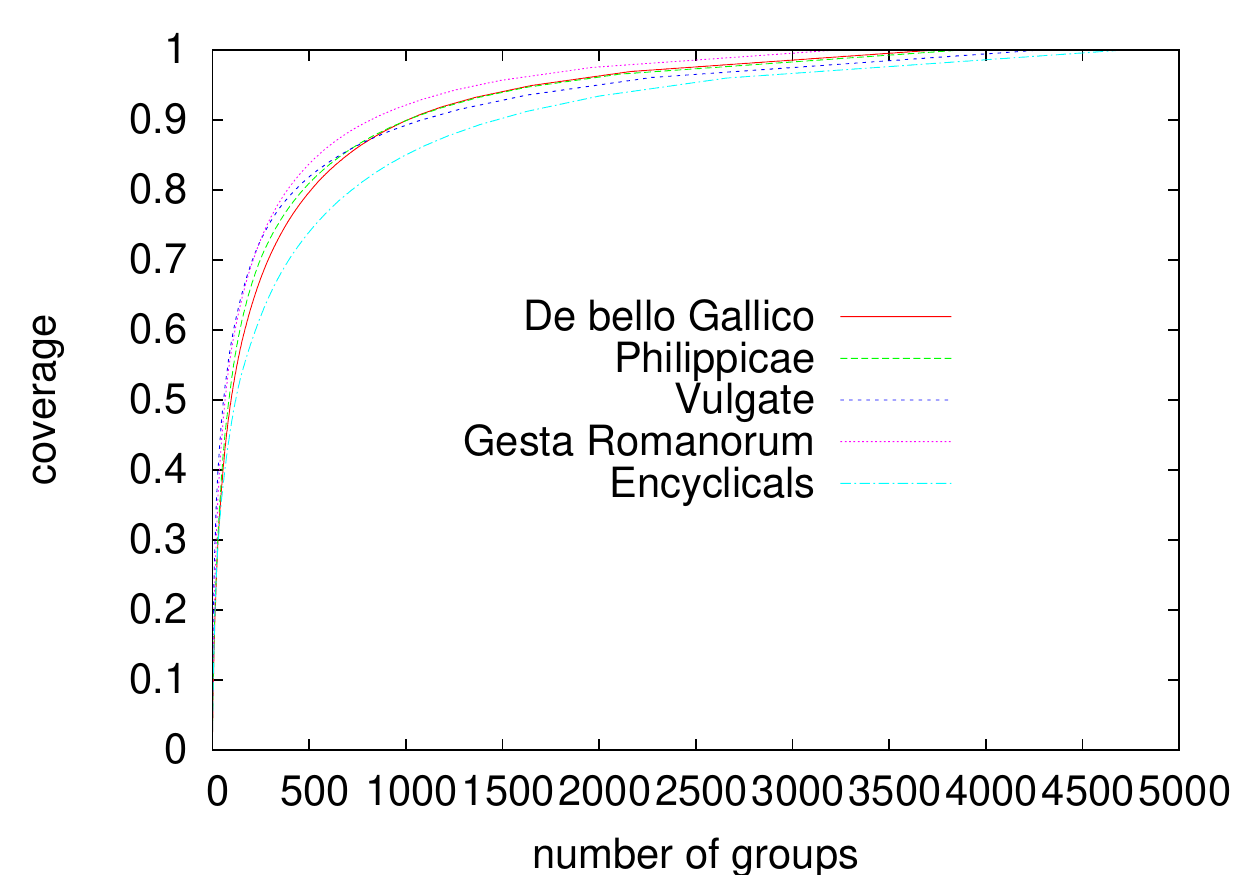}
\caption{Coverage curves for groups.}
\label{covgroups}
\end{figure}
The coverage curves were obtained by the following procedure:
\begin{itemize}
\item The text was converted to lowercase, all punctuation marks
and digits were removed.
\item A list of words was created, together with number of occurrences of each word.
\item For each word in the above list, we used Whitaker's WORD program to
find corresponding headwords.
\item The inflection graph was constructed, and its connected components
determined using Python/NetworkX script.
\item The frequency of occurrence of each word group  was computed,
and coverage curve plotted.
\end{itemize}
Figures~\ref{covwords} and \ref{covgroups} show coverage curves
for words and groups for all five sample texts. 
Comparing these figures 
one can immediately notice two things. First of all, the
coverage converges to $100\%$ faster or slower, depending on the text.
For words coverage, \textit{Vulgate} and \textit{Gesta Romanorum}
are clearly converging faster than both classical Latin texts of Caesar and Cicero and contemporary
encyclicals. This agrees with the general consensus of latinists who consider
medieval texts ``easier'' than classical one.

Another observation is the difference between words coverage and group coverage.
Much smaller number of groups than individual words is needed to achieve 
the same coverage, as one would naturally expect. Table~\ref{covtable}
lists the number of words and groups required to obtain 95\% and 98\%
coverage in all five sample texts. These two numbers have been used because
it is often argued that in order to read a given text with minimal distraction
one needs to know enough words to cover at least  95\%-98\% running words
of the text. Such high coverage is also needed to transfer reading skills
from one's mother tongue to another (foreign or second) language. 
It is encouraging for students of Latin that the number of word groups
required to obtain such coverage is relatively low. Nevertheless,  one has to 
be careful interpreting this table, remembering  that one group may consist of
many dictionary headwords. 

Table~\ref{covtable} also reveals some interesting differences between
classical texts written in literary and vulgar Latin. In order to reach 95\% coverage of \textit{Philipicae} one needs
much larger number of words than in the case of \textit{Vulgate}. On the
other hand,  \textit{Vulgate} requires  larger number of word groups to achieve the same 
coverage. This shows that Cicero used the inflection system of the Latin language
much more skillfully -- he used fewer word
groups than St. Jerome, yet from these he obtained a \textit{larger} number of inflected forms!

\subsection{Normalized coverage}
In order to make coverage curves independent of $N$ and $M$,
one can define the normalized coverage as
\begin{eqnarray}
c_w(x)= C_w(Nx),\\
c_g(x)= C_g(Mx),
\end{eqnarray}
where $x \in [0,1]$.
If the exact form of $n_w(r)$ or $n_g(r)$ was known, the corresponding
coverage could be computed. For example, in the case of the
Zipf law, $n_w(r)=A/r$ where $A$ is a normalization constant.
In such a case, sums required in eq. (\ref{eqcov}) can be
computed in closed forms, yielding
\begin{equation}
c_w(x)=\frac{\Psi(Nx+1)+\gamma}{\Psi(N+1)+\gamma},
\end{equation}
where $\Psi$ is the digamma function, and $\gamma=0.57721566\ldots$ is the  Euler-Mascheroni constant.
\begin{table}
\caption{Table of number of words (w) and groups (g) required to obtain 
95\% and 98\% coverage.}
\label{covtable}
\begin{center}
 \begin{tabular}{l| l | l | l | l}
 & 95\% w & 95\% g & 98\% w & 98\% g \\ \hline
De bello  Gallico& 8415 & 1669 & 9954 & 2714 \\ 
Philipicae & 9172 & 1709 & 10711 & 2850 \\ 
Vulgate & 6937 & 1996 & 8476 & 3261 \\ 
Gesta Romanorum& 6174 & 1378 & 7707 & 2212\\
Encyclicals& 10682 & 2405 & 12221  & 3689
\end{tabular}
\end{center}
\end{table}

\begin{figure}
\centering
\includegraphics[width=3.5in]{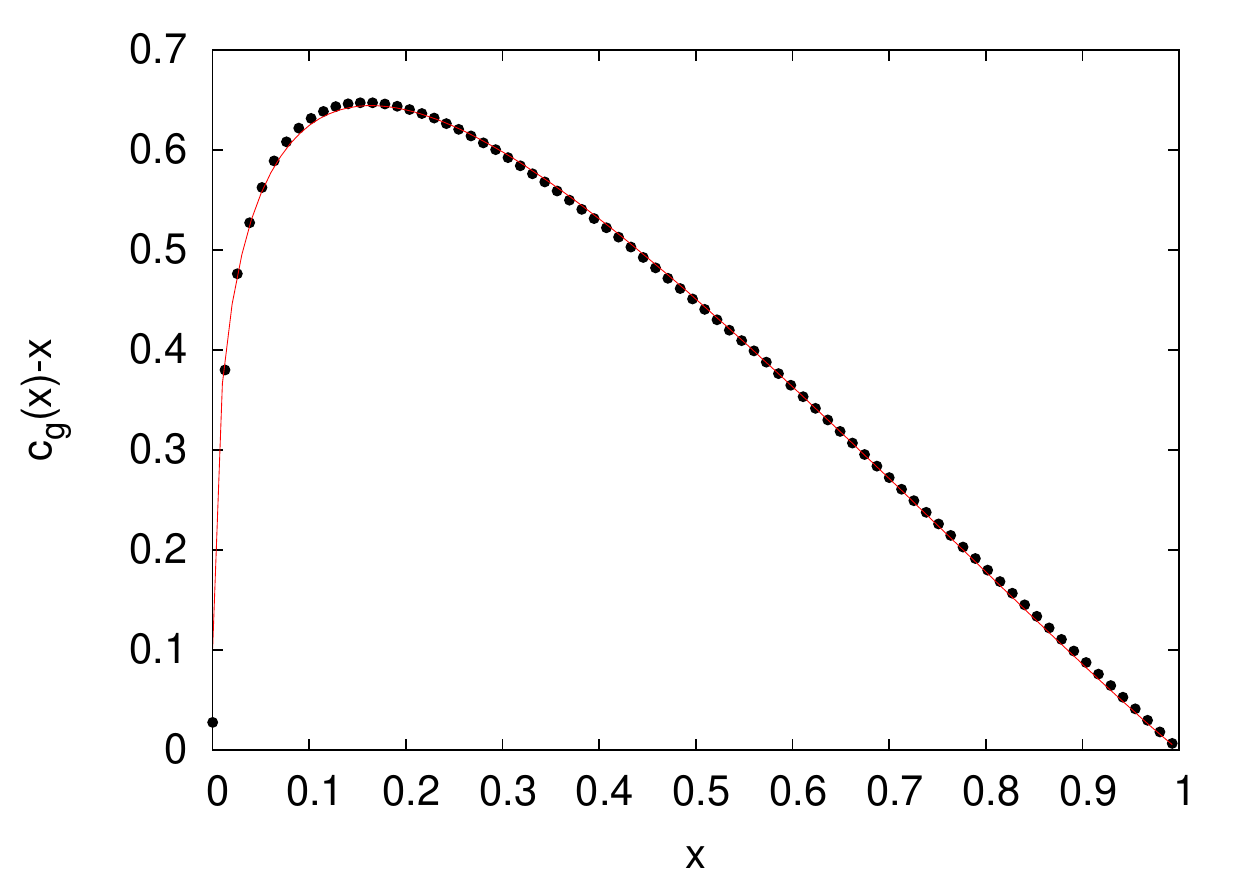}
\caption{Graph of $c_g(x)-x$ as a function of $x$ for
\textit{Encyclicals}. Black dots represent data (every 60 point only plotted), 
and continuous curve is the line of the best fit given by eq. (\ref{fiteq}).}
\label{fitsample}
\end{figure}

Unfortunately, in the case of the group coverage, we do not know what is
the form of $n_g(r)$, thus a similar formula cannot be produced.
It is possible, however, to obtain empirical fit with small number of parameters.
This can be readily understood if we plot $c_g(x)-x$ as a function of $x$,
as shown in Figure~\ref{fitsample}, where, in order to avoid clutter,
only coverage data for one data set (\textit{Encyclicals}) is presented.
Symmetry of this  figure suggests that  $c_g(x)-x$ may be approximated
by the graph of $x^\alpha (1-x)^\beta$, where $\alpha$, $\beta$
are parameters of the fit. 

We, therefore, fitted the function
\begin{equation}
 f(x)=x + x^\alpha (1-x)^\beta
\end{equation}
to the normalized coverage data $c_g(x)$. The fit, although
good, was less than ideal, so we introduced another two parameters,
fitting the function
\begin{equation}\label{fiteq}
 f(x)=x^\gamma + x^\alpha (1-x^\delta)^\beta.
\end{equation}
It should be noted that this choice of the equation does not have any particular meaning, it just has been
observed that it produces a good fit, thus it is a convenient way to describe coverage curves.
Sample fit produced using this equation is shown in Figure~\ref{fitsample}
(continuous line). Values of parameters for all five sample
texts  are shown in table~\ref{fitpars}.
\begin{table}
\caption{Fit parameters}
\label{fitpars}
\begin{center}
 \begin{tabular}{l|l| l|l|l  }
 & $\alpha$  & $\beta$ & $\gamma$ & $\delta$\\ \hline
De bello  Gallico& 0.3688  & 1.3583 &  0.3445  & 0.5877 \\ 
Philipicae & 0.3491  & 1.3707 &  0.3249 &  0.5537 \\ 
Vulgate & 0.3045 &  1.3188  & 0.2781  & 0.4206 \\ 
Gesta Romanorum& 0.3433  & 1.3134 &  0.3196 &  0.5199\\
Encyclicals&  0.3615   &  1.2899  &   0.3407  &   0.5047 
%
\end{tabular}
\end{center}
\end{table}
From the form of eq. (\ref{fiteq}) one can see that the smaller of
parameters $\alpha$ and $\gamma$ controls the initial steepness 
of the normalized coverage curve, and therefore we will define
\begin{equation}
\eta={\mathrm{min}} \{ \alpha, \gamma \}.
\end{equation}
The value of $\eta$  can be given an interpretation related to  the nature of the
underlying text. If $\eta$ is large, it means that the normalized 
coverage curve is growing slowly -- that is, high percentage of 
all word groups present in the text is needed to achieve, say, 95\%
text coverage. On the other hand, if $\eta$ is small, this means steep
coverage curve, so that high coverage is reached quickly. 

For that reason, one can say that the parameter $\eta$ tells us to what
degree is the the inflection mechanism of the language used in order 
to provide high text coverage. Small $\eta$ means high reliance
on the inflection mechanism. From Table~\ref{fitpars} we can therefore
conclude that  \textit{Vulgate} and \textit{Gesta Romanorum} do not rely on the inflection
as much as the classical works or  modern encyclicals.

\section{Inflection graph for a dictionary}
In previous sections, we were considering inflection graphs for individual
literary works. It is possible to obtain such graph for the whole Latin 
language -- or, to be more precise, for all words from a large dictionary.
In Whitaker's WORDS dictionary, there are 35,670 distinct
headwords. Using WORDS program, one can construct all possible inflected
forms for all these headwords, resulting in a list of 1,032,669 word forms.
We shall note that those are all \textit{theoretically} possible forms,
and that not all of them are attested in the surviving corpus of Latin
texts. For example, the form \textit{coquor} is a theoretically possible
passive, present tense, first person singular of \textit{coquo} (to cook),
yet in does not seem to be attested in Latin texts~\cite{verbs501}. 

If we connect all headwords from the dictionary with all theoretically
possible inflected forms, we obtain an inflection graph of the whole language.
The resulting graph is very sparse, having 1,117,394 edges, that is, only 
10\% more than the number of edges. The reason fort his is that the vast majority 
of inflected forms correspond to only one dictionary headword. 
\begin{figure}[!t]
\centering
\includegraphics[width=3.5in]{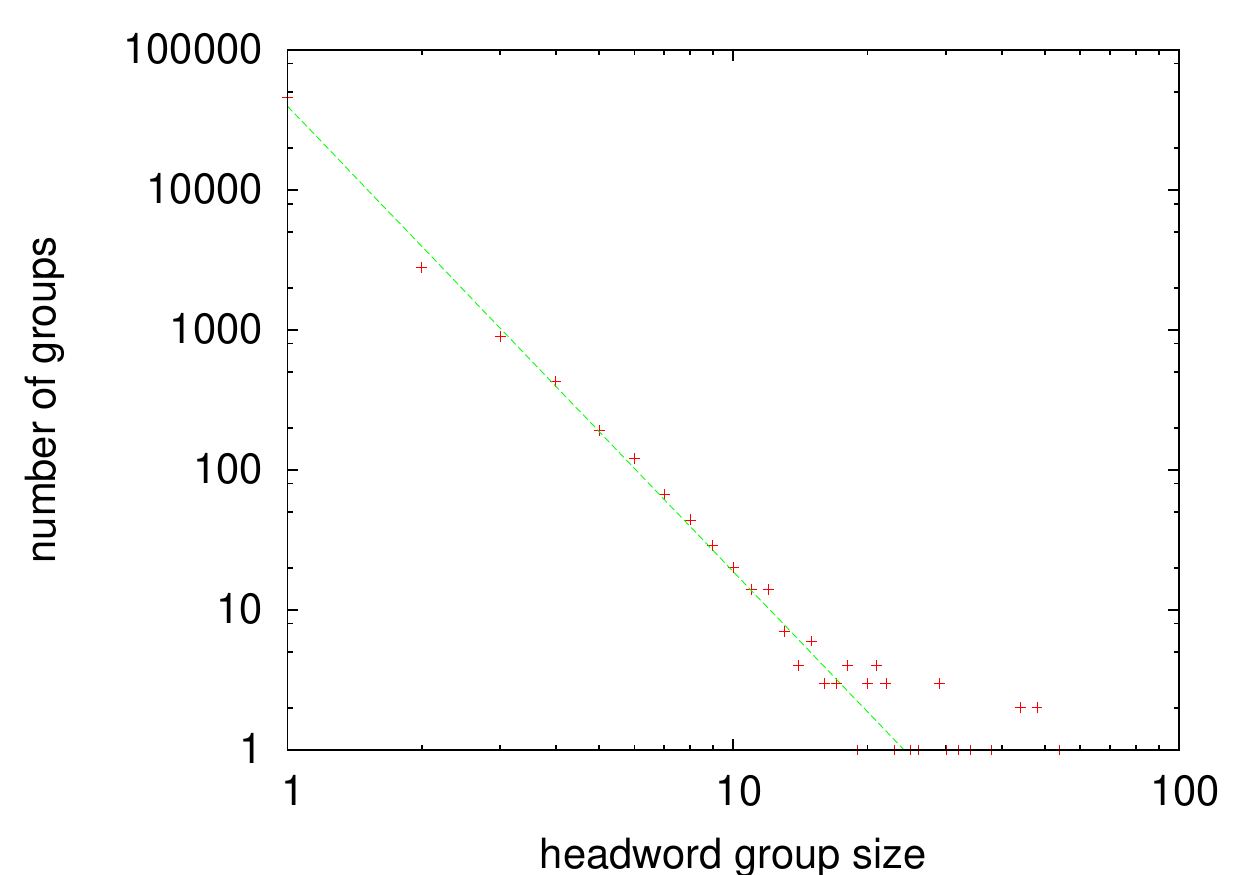}
\caption{Distribution of sizes of headword groups in all Latin words from Whitaker's WORDS dictionary.}
\label{clheadstats}
\end{figure}
The number of connected components of this graph is 50,847, and their size
distribution appears to follow a power law. This is illustrated
in Figure~\ref{clheadstats}. If $H(m)$ is the number of 
groups with $m$ headwords, then the line of the best
fit shown in  Figure~\ref{clheadstats} closely follows
the power law
\begin{equation}
 H(m) \sim m^{-\tau},
\end{equation}
where the  value of the exponent $\tau$ obtained by fitting a straight line to 
datapoints excluding several largest clusters is $\tau=3.32$.

This phenomenon strongly resembles percolation.
The scaling theory for percolation predicts that connected
components exhibit close to power-law behavior near the percolation threshold.
This seems to suggest that the sparse inflection graph discussed here
may be close to its percolation threshold. 

Let us now make some remarks regarding  largest components,
corresponding to data points lying above the fitted line on the right
of Figure~\ref{clheadstats}.
This deviation could be caused by a finite size of the graph. Similar 
behavior is often observed in numerical simulations of percolation
in finite systems.

\begin{figure}
\centering
\includegraphics[width=3.5in]{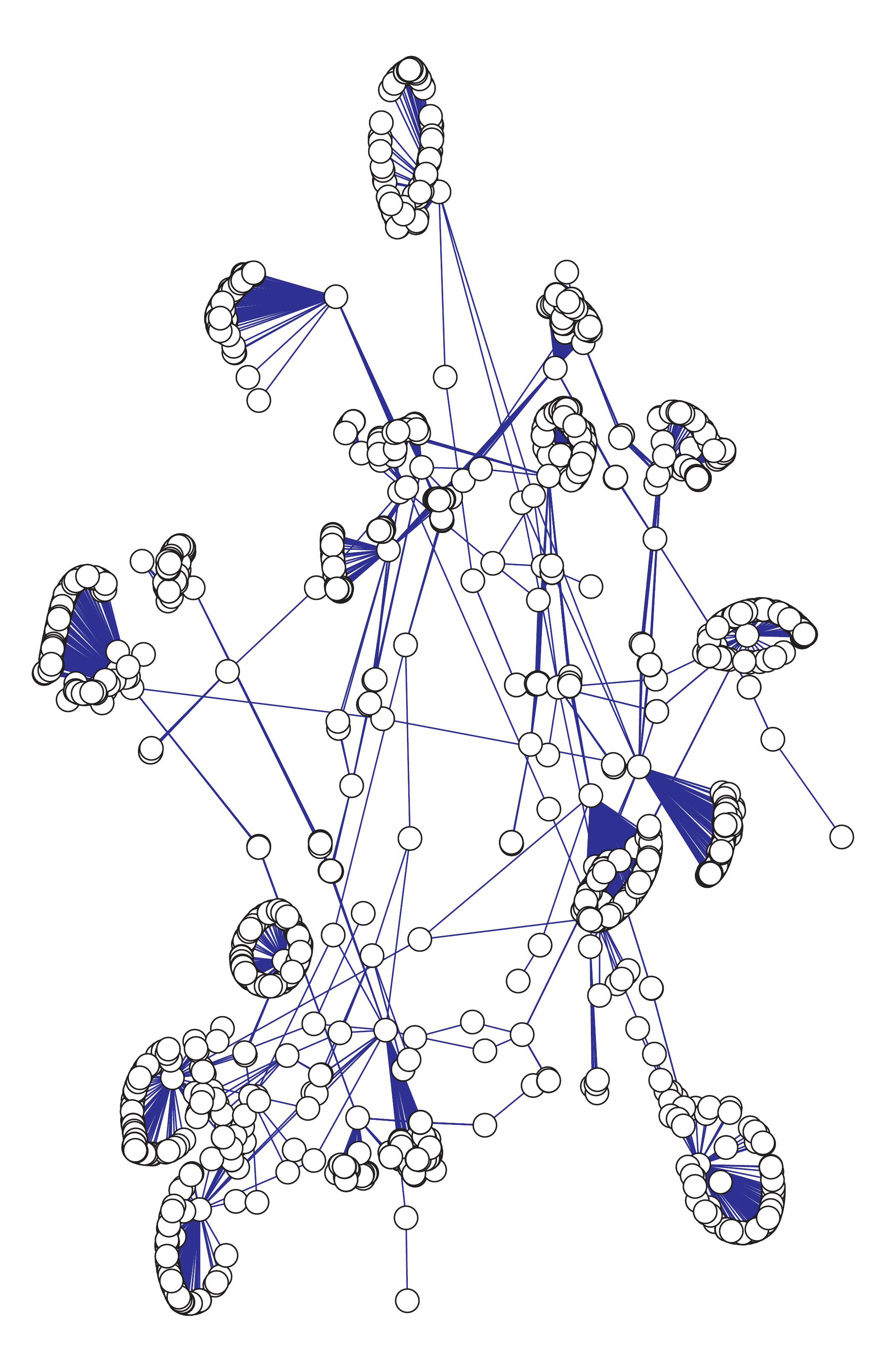}
\caption{Visualization of the largest component of the dictionary inflection graph.}
\label{bigcluster}
\end{figure}

Obviously,
the headwords belonging to very large clusters cannot be all semantically related, and 
the fact that they are grouped together may be  somewhat related to the inclusion of
all theoretically possible inflected forms. A path
 joining two vertices may exist solely because it passes through
some inflected form which is unattested. In Figure~\ref{bigcluster} the largest cluster
of the dictionary inflection graph is shown. One can see from this picture that
the cluster is composed of several one-level trees (stars) loosely connected via
a number of bridges.  Some of these bridges are likely 
``artificial'' (unattested) forms, and removing them  would most likely divide the big cluster 
into a number of smaller clusters. 

\section{Conclusion}
We presented some preliminary findings regarding properties of inflection graphs. 
The concept of the word group defined as a connected component of the inflection
graph appears to be useful in describing the vocabulary structure of the text.
The parameter $\eta$ could be used to characterize some aspects of the text difficulty,
describing the balance between the diversity of vocabulary versus the diversity of inflected forms.
Obviously, if one wants to categorize texts according to their perceived difficulty, 
vocabulary is not the only factor. Structure of sentences and word order are 
equally important, or sometimes even more important, thus such categorization
scheme will most likely involve several parameters. The author hopes that
an automated system classifying Latin texts is eventually constructed, since such
system would be very beneficial to students of Latin. In languages such as English,
sets of graduated books exist, and are routinely used in classrooms. Students start
from very simple texts and seamlessly progress to texts of increasing difficulty.
In Latin it is very unlikely that such set of books would be ever written, given the
current status of the language. Nevertheless, texts of various levels of difficulty
do exist in Latin -- yet it is often hard for beginners to determine beforehand which
texts (out of thousands available) should be read first, and which should be left
till later. We plan to develop ideas presented here to create an automated system which
could be used to ``measure'' text difficulty and eventually mine internet repositories
to produce a list of Latin text forming a sequence with gradually increasing difficulty.

The similarity of the dictionary inflection graph to percolating network  is also currently 
investigated. We are collecting a large corpus of Latin texts in order to remove from the
graph all unattested words. Due to variation of spelling, especially in Medieval texts,
this task cannot be fully automated, thus  final results are not yet  available. We also
plan to find out whether other synthetic languages exhibit similar scaling 
phenomenon as the Latin inflection graph.


\section*{Acknowledgment}
The author acknowledges partial financial support from the Natural Sciences and Engineering Research Council of Canada
in the form of Discovery Grant.






%


\begin{thebibliography}{10}
\providecommand{\url}[1]{#1}
\csname url@samestyle\endcsname
\providecommand{\newblock}{\relax}
\providecommand{\bibinfo}[2]{#2}
\providecommand{\BIBentrySTDinterwordspacing}{\spaceskip=0pt\relax}
\providecommand{\BIBentryALTinterwordstretchfactor}{4}
\providecommand{\BIBentryALTinterwordspacing}{\spaceskip=\fontdimen2\font plus
\BIBentryALTinterwordstretchfactor\fontdimen3\font minus
  \fontdimen4\font\relax}
\providecommand{\BIBforeignlanguage}[2]{{%
\expandafter\ifx\csname l@#1\endcsname\relax
\typeout{** WARNING: IEEEtran.bst: No hyphenation pattern has been}%
\typeout{** loaded for the language `#1'. Using the pattern for}%
\typeout{** the default language instead.}%
\else
\language=\csname l@#1\endcsname
\fi
#2}}
\providecommand{\BIBdecl}{\relax}
\BIBdecl

\bibitem{Ferrer2001}
R.~F. i~Cancho1 and R.~V. Sol\'s, ``The small world of human language,''
  \emph{Proc. Roy. Soc. Lond. B}, vol. 268, pp. 2261--2265, 2001.

\bibitem{MotterdLD02}
A.~E. Motter, A.~P.~S. de~Moura, Y.~C. Lai, and P.~Dasgupta, ``Topology of the
  conceptual network of language,'' \emph{Phys. Rev. E}, vol.~65, 2002.

\bibitem{KinouchiMLLR02}
O.~Kinouchi, A.~S. Martinez, G.~F. Lima, G.~M. Lourenco, and S.~Risau-Gusman,
  ``Deterministic walks in random networks: an application to thesaurus
  graphs,'' \emph{Physica A}, vol. 315, pp. 665--676, 2002.

\bibitem{HolandaPKMR04}
A.~D. Holanda, I.~T. Pisa, O.~Kinouchi, A.~S. Martinez, and E.~E.~S. Ruiz,
  ``Thesaurus as a complex network,'' \emph{Physica A-Statistical Mechanics And
  Its Applications}, vol. 344, pp. 530--536, 2004.

\bibitem{citeulike:1179006}
\BIBentryALTinterwordspacing
M.~Sigman and G.~A. Cecchi, ``Global organization of the wordnet lexicon,''
  \emph{PNAS}, vol.~99, no.~3, pp. 1742--1747, February 2002. [Online].
  Available: \url{http://dx.doi.org/10.1073/pnas.022341799}
\BIBentrySTDinterwordspacing

\bibitem{KeY08}
J.~Y. Ke and Y.~Yao, ``Analysing language development from a network
  approach,'' \emph{Journal Of Quantitative Linguistics}, vol.~15, pp. 70--99,
  2008.

\bibitem{CaldeiraLANM06}
S.~M.~G. Caldeira, T.~C.~P. Lobao, R.~F.~S. Andrade, A.~Neme, and J.~G.~V.
  Miranda, ``The network of concepts in written texts,'' \emph{European
  Physical Journal B}, vol.~49, pp. 523--529, 2006.

\bibitem{PomiM04}
A.~Pomi and E.~Mizraji, ``Semantic graphs and associative memories,''
  \emph{Physical Review E}, vol.~70, p. 066136, 2004.

\bibitem{CanchoSK04}
R.~F.~I. Cancho, R.~V. Sole, and R.~Kohler, ``Patterns in syntactic dependency
  networks,'' \emph{Physical Review E}, vol.~69, p. 051915, 2004.

\bibitem{AntiqueiraNOC07}
L.~Antiqueira, M.~G.~V. Nunes, O.~N. Oliveira, and L.~D. Costa, ``Strong
  correlations between text quality and complex networks features,''
  \emph{Physica A-Statistical Mechanics And Its Applications}, vol. 373, pp.
  811--820, 2007.

\bibitem{words}
{W. Whitaker}, ``{WORDS}, {L}atin-{E}nglish dictionary,''
  http://users.erols.com/whitaker/words.htm.

\bibitem{Diederich39}
\BIBentryALTinterwordspacing
P.~B. Diederich, ``The frequency of {L}atin words and their endings,'' Ph.D.
  dissertation, Columbia University, 1939. [Online]. Available:
  \url{http://users.erols.com/whitaker/freq.htm}
\BIBentrySTDinterwordspacing

\bibitem{walrus}
{CAIDA}, ``Walrus -- graph visualization tool,''
  http://www.caida.org/tools/visualization/walrus.

\bibitem{Networkx}
``Network{X}, {P}ython package for analysis of complex networks,''
  https://networkx.lanl.gov.

\bibitem{vatican}
{Ioannes Paulus PP. II}, ``Litterae encyclicae,'' http://www.vatican.va.

\bibitem{latlib}
``The latin library,'' http://www.thelatinlibrary.com.

\bibitem{verbs501}
R.~E. Prior and J.~Wohlberg, \emph{501 Latinn Verbs}.\hskip 1em plus 0.5em
  minus 0.4em\relax Hauppauge, NY: Barron's, 1995.

\end{thebibliography}
\end{document}